\documentclass[twoside,11pt]{article}

\usepackage[preprint]{dmlr2e}
\usepackage{etoolbox}
\makeatletter
\patchcmd{\@maketitle}{\\ \\ {\bf Reviewed on OpenReview:} \openreview}{}{}{}
\makeatother
\usepackage[utf8]{inputenc} %
\usepackage[T1]{fontenc}    %
\usepackage{url}            %
\usepackage{booktabs}       %
\usepackage{textcomp}       %
\usepackage{amsfonts}       %
\usepackage{microtype}      %
\usepackage{xcolor}         %
\usepackage{amsmath}
\usepackage{lastpage}

\newcommand{\sd}[1]{\,{\scriptsize(\textpm#1)}}

\usepackage{tikz}
\usepackage{graphicx}
\usepackage{worldflags}

\DeclareRobustCommand{\globe}{%
  \begin{tikzpicture}[baseline=-0.6ex, line width=0.5pt, x=0.165cm, y=0.165cm]
    \draw (0,0) circle (1);
    \draw (0,0) ellipse (0.42 and 1);
    \draw (0,-1) -- (0,1);
    \draw (-1,0) -- (1,0);
    \draw (-0.87,0.5) -- (0.87,0.5);
    \draw (-0.87,-0.5) -- (0.87,-0.5);
  \end{tikzpicture}}

\definecolor{barcol}{HTML}{3B6EA8}
\definecolor{barours}{HTML}{C0392B}
\def\resW{2.4}                 %
\def\resLo{1}\def\resHi{100}   %
\newcommand{\resx}[1]{{ln(#1/\resLo)/ln(\resHi/\resLo)*\resW}}
\newcommand{\resbar}[4][barcol]{%
  \begin{tikzpicture}[baseline=-0.6ex]
    \useasboundingbox (-2pt,-4pt) rectangle ([xshift=1.3cm]\resW,4pt);
    \draw[black!12, line width=2.4pt] (0,0) -- (\resW,0);
    \foreach \t in {1,10,100} \draw[black!25] (\resx{\t},-2.2pt) -- (\resx{\t},2.2pt);
    \ifdim #2pt=#3pt
      \fill[#1] (\resx{#2},0) circle (2pt);
    \else
      \draw[#1, line width=3.4pt] (\resx{#2},0) -- (\resx{#3},0);
    \fi
    \node[anchor=west, font=\scriptsize, text=black!60, inner sep=0pt]
      at ([xshift=3pt]\resW,0) {#4};
  \end{tikzpicture}}
\newcommand{\resaxis}{%
  \begin{tikzpicture}[baseline=-0.6ex, font=\tiny, text=black!55]
    \useasboundingbox (-2pt,-3pt) rectangle ([xshift=1.3cm]\resW,3pt);
    \foreach \t/\l in {1/1\,cm,10/10\,cm,100/100\,cm} \node[inner sep=0pt] at (\resx{\t},0) {\l};
  \end{tikzpicture}}

\newcommand{\flag}[1]{\hbox{\worldflag[length=0.42cm, width=0.28cm, framewidth=0.1pt]{#1}\unskip}}

\newsavebox{\fitbox}
\newcommand{\fitwidth}[1]{%
  \sbox{\fitbox}{#1}%
  \ifdim\wd\fitbox>\linewidth
    \resizebox{\linewidth}{!}{\usebox{\fitbox}}%
  \else
    \usebox{\fitbox}%
  \fi}

\dmlrheading{XX}{2026}{1-\pageref{LastPage}}{MM/26}{MM/26}{XX-XXXX}{Ayushi Sharma, Clemens Mosig, Lukas Drees, Salim Soltani, Janusch Vajna-Jehle, Aaron Sheppard, Belqis Ahmadi, Jonathan Schmid, Paul Neumeier, Nathan Jacobs, Jan Dirk Wegner, and Teja Kattenborn}
\def\openreview{\url{https://openreview.net/forum?id=XXXX}} %

\ShortHeadings{DTE-aerial: Tree Cover and Mortality Dataset}{Sharma et al.}
\firstpageno{1}

\title{deadtrees.earth-aerial: A Multi-Resolution Aerial Image Dataset for Tree Cover and Mortality Detection}

\author{\name Ayushi Sharma \email ayushi.sharma@geosense.uni-freiburg.de \\
       \addr Chair of Sensor-based Geoinformatics, University of Freiburg, Germany\\
       \addr EcoVision Lab, DM3L, University of Zurich, Switzerland
       \AND
       \name Clemens Mosig \email clemens.mosig@uni-leipzig.de \\
       \addr Institute for Earth System Science and Remote Sensing, Leipzig University, Germany
       \AND
       \name Lukas Drees \\
       \addr EcoVision Lab, DM3L, University of Zurich, Switzerland
       \AND
       \name Salim Soltani \\
       \name Janusch Vajna-Jehle \\
       \name Aaron Sheppard \\
       \name Belqis Ahmadi \\
       \name Jonathan Schmid \\
       \addr Chair of Sensor-based Geoinformatics, University of Freiburg, Germany
       \AND
       \name Paul Neumeier \\
       \addr Institute for Earth System Science and Remote Sensing, Leipzig University, Germany
       \AND
       \name Nathan Jacobs \\
       \addr Washington University in St. Louis, USA
       \AND
       \name Jan Dirk Wegner \\
       \addr EcoVision Lab, DM3L, University of Zurich, Switzerland
       \AND
       \name Teja Kattenborn \email teja.kattenborn@geosense.uni-freiburg.de \\
       \addr Chair of Sensor-based Geoinformatics, University of Freiburg, Germany}

\editor{TBD}

\begin{document}

\maketitle

\begin{abstract}%
Forests worldwide are increasingly threatened by climate change and disturbances such as fire, pests, and pathogens, creating an urgent need for scalable monitoring of tree cover and tree mortality.
Aerial imagery from drones and aircraft is a key data source for detailed and large-scale mapping of tree crowns and mortality. However, progress is limited by the lack of globally distributed, multi-biome, harmonized datasets for joint segmentation of tree cover and mortality. 
To address this, we introduce two novel, open, machine-learning-ready datasets to enable joint segmentation of tree cover and tree mortality from centimeter-scale aerial imagery across globally distributed sites.
Our training dataset, DTE-aerial-train, comprises 346K image patches of size $1024 \times 1024$ pixels, with spatial resolutions ranging from 2.5 to 20\,cm.
It includes audited pseudo-labels for tree cover and mortality across the training set, with expert annotations for rare mortality sites.
Our evaluation dataset, DTE-aerial-bench, provides a multi-biome, multi-resolution benchmark test set of 25 globally distributed orthoimages totaling 525 patches with high-quality expert annotations for both tree cover and mortality.
Both the training and benchmark datasets span tropical, temperate, boreal, and dryland biomes and cover a wide range of forest structures and mortality patterns.
Using the benchmark set, we establish reference baselines for mortality and tree cover segmentation across diverse biomes and spatial resolutions. Improvements over prior models vary by biome: they are largest in boreal and montane forests (F1 from 0.40 to 0.58), and modest in tropical forests and drylands.
Across three spatial resolutions, the new model improved estimates of site-level mortality fractions, increasing mean \(R^2\) from 0.377 to 0.917 and reducing mean absolute error by 49\% relative to the baseline.

For data, model and code, please visit our \href{https://ayushi-3536.github.io/deadtrees_aerial.github.io/}{project page}.
An interactive visualization of the benchmark dataset is also available at \href{https://deadtrees.earth/releases/dte-aerial-bench}{our release page}.

\end{abstract}

\begin{keywords}
  tree mortality, tree cover, semantic segmentation, aerial imagery, remote sensing, benchmark dataset, multi-resolution
\end{keywords}

\section{Introduction}

The world's forests shape carbon, water, and energy fluxes while providing essential ecosystem services for human well-being, including carbon storage, biodiversity habitat, water provisioning, and protection from natural hazards \citep{perry2008forest, friedlingstein2023global}.
Under ongoing climate change, forests worldwide are increasingly exposed to rising atmospheric aridity and more frequent and prolonged droughts, leading to elevated tree mortality \citep{hartmann2022climate, trumbore2015forest, allen2015underestimation}. %
These impacts are further exacerbated by pests, pathogens, forest degradation, and wildfires \citep{scholze2006climate, raffa2008cross, bachelet2003simulating}. %
Understanding how these stressors shape forest dynamics worldwide requires globally consistent monitoring of tree cover and mortality \citep{migliavacca2025securing, international2025towards}.

Forest monitoring in this context requires accurate mapping of tree crown cover across spatial scales, as crown cover changes reflect processes such as recovery, disturbance, and degradation. In parallel, it requires precise detection of mortality in standing trees ranging from early-stage signals such as partial crown or branch dieback to fully dead standing tree crowns \citep{international2025towards, mosig2026deadtrees}.

Aerial imagery from aircraft and drones is a key data source for forest monitoring, bridging the gap between coarse-resolution satellite data and spatially limited terrestrial measurements \citep{schiefer2023uav}.
Through national airborne campaigns, aerial imagery from aircraft has become widely available, while locally acquired drone imagery has also become abundant, emerging as a key technology for environmental monitoring \citep{kattenborn2021review, mosig2026deadtrees}.

Airplane and drone imagery combined with modern computer vision enables scalable extraction of tree cover and tree mortality \citep{mosig2026deadtrees, mohring2025global, schiefer2023uav, veitch2024oam}. In particular, centimeter-scale aerial RGB imagery ($<$10\,cm) provides the spatial detail and availability needed for this task,
often outperforming more complex modalities such as multispectral or hyperspectral imagery \citep{khatri2024enhancing, kattenborn2021review}. While prior work has shown strong performance in both tree cover mapping and mortality detection \citep{veitch2024oam, mohring2025global, schiefer2023uav}, these tasks have so far only been addressed in isolation. 

Delineating both classes jointly is however key to be able to compute mortality as a fraction of the tree cover in which it occurs.
OAM-TCD \citep{veitch2024oam} provides tree-cover annotations for 541 orthoimages worldwide, but contains no mortality labels and does not delineate canopy gaps in closed canopies. Conversely, the deadtrees.earth mortality dataset \citep{mohring2025global} provides mortality annotations for 434 globally distributed sites, but does not delineate living tree cover and under-represents brown-stage mortality such as fire. 

\begin{figure}[tp]
    \centering
    \includegraphics[width=\columnwidth]{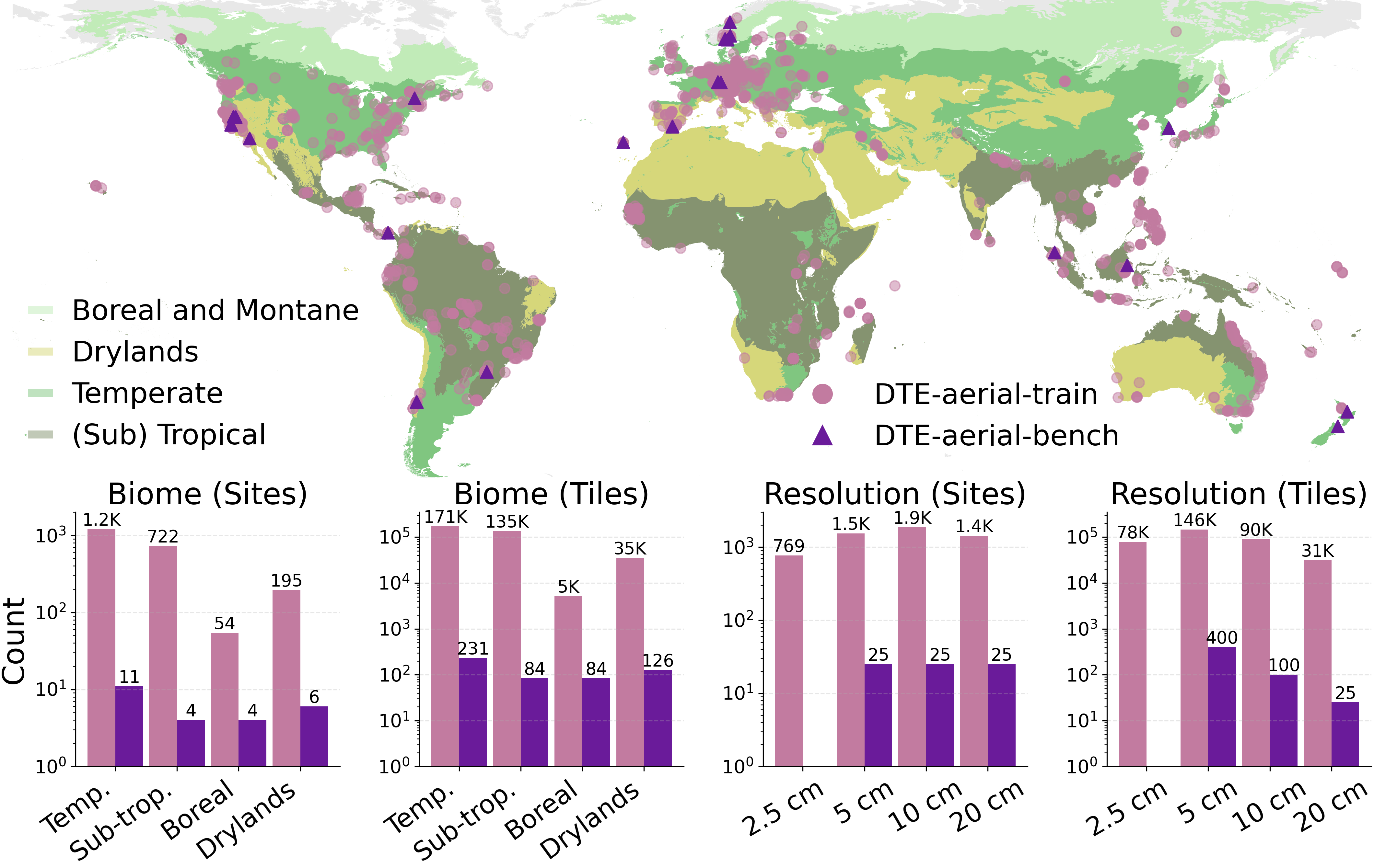}
    \caption{
    \textbf{Dataset overview.}
    (Top) Global, crowd-sourced aerial imagery from \textit{deadtrees.earth}, representing one of the largest high-resolution collections for tree mortality. The training set spans tropical, temperate, boreal, and dryland biomes, capturing diverse canopy structures and forest conditions, while the benchmark is a curated subset emphasizing label quality. 
    (Bottom) Distribution of sites and patches across biome groups and spatial resolutions for training and benchmark sets. Patch counts for the training set are reported in thousands to reflect scale.
    }
    \label{fig:dataset_overview}
\end{figure}

Existing annotated aerial image datasets of forests further do not comprehensively represent the diversity of forest types, species compositions, and functional types across global biomes \citep{mosig2026deadtrees}. In addition, aerial images further differ substantially in spatial resolution, sensor characteristics, and illumination conditions, which can affect the spatial features learned by models and limit their generalization across acquisition conditions \citep{mohring2025global, schiefer2024large}. Together, these factors make it difficult to develop and evaluate models across diverse forest environments.

These gaps motivate a benchmark with four properties: (1) joint, coherent labels for tree cover and mortality within the same scenes, enabling mortality to be expressed as a fraction of tree covered area; (2) coverage of diverse mortality scenarios, including early dieback or fire disturbance; (3) cross-scale consistency, so that spatial resolution is the only varying factor when the same scene is evaluated at different scales; and (4) expert annotation with iterative consensus and multi-reviewer approval

With the support of deadtrees.earth (DTE) initiative, we make the following contributions:
\begin{itemize}
    \item We enable standardized evaluation of tree cover and mortality segmentation across biomes and spatial resolutions through \textbf{DTE-aerial-bench}: a multi-resolution, globally distributed, multi-biome benchmark dataset with expert-annotated, cross-scale consistent of tree cover and standing deadwood.
    \item We introduce \textbf{DTE-aerial-train}, a globally distributed, centimeter-scale, multi-resolution labeled aerial image dataset covering 2,176 sites for jointly mapping tree cover and tree mortality.
    \item We establish \textbf{baseline models} for simultaneous multi-class segmentation of tree cover and mortality across diverse biomes and image resolutions, providing initial benchmarks and highlighting challenges in generalization under geographic and multi-resolution variability, as well as severe class imbalance between abundant tree cover and rare mortality instances. Beyond pixel-level metrics, we evaluate site-level mortality fractions and assess robustness with leave-one-site-out analyses.
\end{itemize}

\begin{table}[tp]
\centering
\caption{Comparison of existing open and ML-ready aerial image datasets and benchmarks for tree cover and mortality mapping. With DTE-aerial-train and DTE-aerial-bench, we provide the first centimeter-scale and multi-resolution datasets with consistent semantic segmentation annotations for both tree cover and mortality, offering global coverage across the largest quantity of high-resolution orthoimagery (sites). Resolution bars show the ground sampling distance on a log scale (dots mark a single resolution). Coverage: flags denote the countries covered, \globe{} denotes global coverage.}
\label{tab:dataset_comparison}
\small
\fitwidth{%
\begin{tabular}{lcclrc}
\toprule
\textbf{Dataset} &
\textbf{Tree Cover} &
\textbf{Mortality} &
\multicolumn{1}{c}{\textbf{Resolution}} &
\textbf{Sites} &
\textbf{Coverage} \\
 & & & \resaxis & & \\[-5pt]
\midrule
BAMForest~\citep{troles2024bamforests} & \checkmark & -- & \resbar{10}{20}{10--20} & 4 & \flag{DE} \\
SelvaMask~\citep{duguay2026selvamask} & \checkmark & -- & \resbar{1}{4}{1--4} & 3 & \flag{PA}\hspace{2pt}\flag{BR}\hspace{2pt}\flag{EC} \\
\citet{ball2023accurate} & \checkmark & -- & \resbar{8}{10}{8--10} & 4 & \flag{MY}\hspace{2pt}\flag{GF} \\
OAM-TCD~\citep{veitch2024oam} & \checkmark & -- & \resbar{10}{10}{10} & 541 & \globe \\
TreeFinder~\citep{wangtreefinder} & -- & \checkmark & \resbar{60}{60}{60} & 1,000 & \flag{US} \\
\citet{mohring2025global} & -- & \checkmark & \resbar{1}{28}{1--28} & 434 & \globe \\
FORTRESS~\citep{schiefer2022fortress} & \checkmark & \checkmark & \resbar{1}{2}{1--2} & 51 & \flag{DE} \\
\citet{allen2024low} & \checkmark & \checkmark & \resbar{3}{3}{3} & 9 & \flag{ES} \\
\midrule
\textbf{DTE-aerial-train (ours)} & \checkmark & \checkmark & \resbar[barours]{2.5}{20}{\textbf{2.5--20}} & \textbf{2,176} & \globe \\
\textbf{DTE-aerial-bench (ours)} & \checkmark & \checkmark & \resbar[barours]{5}{25}{\textbf{5--25}} & \textbf{25} & \globe \\
\bottomrule
\end{tabular}%
}
\end{table}

\section{Related work}
\label{relatedwork}

While a range of methods, tasks and datasets have been proposed for tree and tree-mortality monitoring from aerial imagery \citep{kattenborn2021review}, previous research showed that detecting partial crown decline and small canopy gaps requires fine-grained delineation \citep{kattenborn2021review, schiefer2023uav, mohring2025global}. %
In contrast, tree instance segmentation and object detection~\citep{weinstein2020training, polewski2015detection, cheng2024scattered} are often ill-suited to many forest ecosystems where overlapping and interwoven crowns diffuse tree boundaries~\citep{allen2025manual, kattenborn2021review}.%
We therefore focus on \emph{semantic segmentation}, as pixel-level information and exact shapes are essential for accurately quantifying tree cover and mortality. Previous work has largely focused on either semantic segmentation of tree cover \emph{or} tree mortality, but jointly only at national scale (Table~\ref{tab:dataset_comparison}).
We therefore review datasets and baselines for each task separately before positioning our contributions.

\paragraph{Previous works on tree cover estimation.}

Computer vision has been widely applied to tree crown segmentation from remote sensing imagery~\citep{kattenborn2021review}. High-quality datasets and corresponding model developments are available for temperate forests~\citep{cloutier2024influence, troles2024bamforests, schiefer2020mapping} as well as tropical forests~\citep{duguay2026selvamask, ball2023accurate, vasquez2023barro}. While these datasets have enabled strong methodological progress, they are typically developed and evaluated on geographically constrained study areas, limiting generalizability of the resulting models across the world's forest ecosystems (Table~\ref{tab:dataset_comparison}, \citet{kattenborn2022spatially}).

A major step toward broader coverage was made by OAM-TCD~\citep{veitch2024oam}, which provides tree-cover annotations for 541 drone orthoimages worldwide. However, the dataset is restricted to 10\,cm spatial resolution and contains relatively weak annotations in dense canopies, limiting the ability of models to learn fine canopy structures such as narrow canopy gaps in closed forests~\citep{veitch2024oam, mosig2026deadtrees, mosig2026sub}. More generally, current tree cover datasets do not provide coherent joint labels for both tree cover and mortality, which are required for consistent forest vitality assessment~\citep{international2025towards}.

\paragraph{Previous works on tree mortality.}
Recent work has also successfully demonstrated the detection of tree mortality and crown decline (often referred to as standing deadwood) in aerial imagery; in this work, we use the term ``mortality'' to encompass both. 

For example, \citet{allen2024low} introduced a labeled drone image dataset at 3\,cm spatial resolution covering approximately 1,500\,ha of \emph{Pinus pinea} forest across nine sites in Pinar de Almorox, Spain. FORTRESS~\citep{schiefer2022fortress} provides 51 labeled orthomosaics from mixed temperate forests in Germany at 1--2\,cm resolution. These datasets demonstrate the value of centimeter-scale imagery for detecting fine-grained mortality patterns, but are limited in geographic and ecological scope.

More recently, larger-scale initiatives have emerged. TreeFinder~\citep{wangtreefinder} aggregates tree-mortality data at 60\,cm spatial resolution from more than 1,000 sites, but is limited to the contiguous United States. In addition, 60\,cm resolution is too coarse to reliably detect deciduous dead trees, partial dieback, and other fine-scale mortality signals~\citep{cheng2024scattered, mosig2026deadtrees, schiefer2023uav, kattenborn2021review}. Thus, although these datasets are valuable contributions, they remain limited for large-scale forest monitoring in their combined coverage of geographic regions, disturbance types, and fine-scale mortality expression.

The most geographically comprehensive mortality-focused dataset to date was collected through the deadtrees.earth platform (\url{https://deadtrees.earth}), an initiative designed to crowdsource high-resolution aerial imagery for tree-mortality mapping \citep{mohring2025global}. The resulting open dataset contains fine-resolution drone imagery (0.01--28\,cm) from 434 sites worldwide and enabled training of a globally transferable mortality model \citep{mohring2025global}. However, it lacks integrated labels for jointly modeling living tree cover and mortality, preventing coherent estimation of relative mortality fraction within forested areas, and it is largely biased toward temperate forests. %

Overall, existing datasets reflect a trade-off between spatial resolution, annotation quality, geographic coverage, and label completeness. There is still no globally distributed, centimeter-scale, multi-resolution dataset with coherent annotations for both tree cover and mortality.%

\begin{figure}[tp]
    \centering
    \includegraphics[width=0.85\columnwidth]{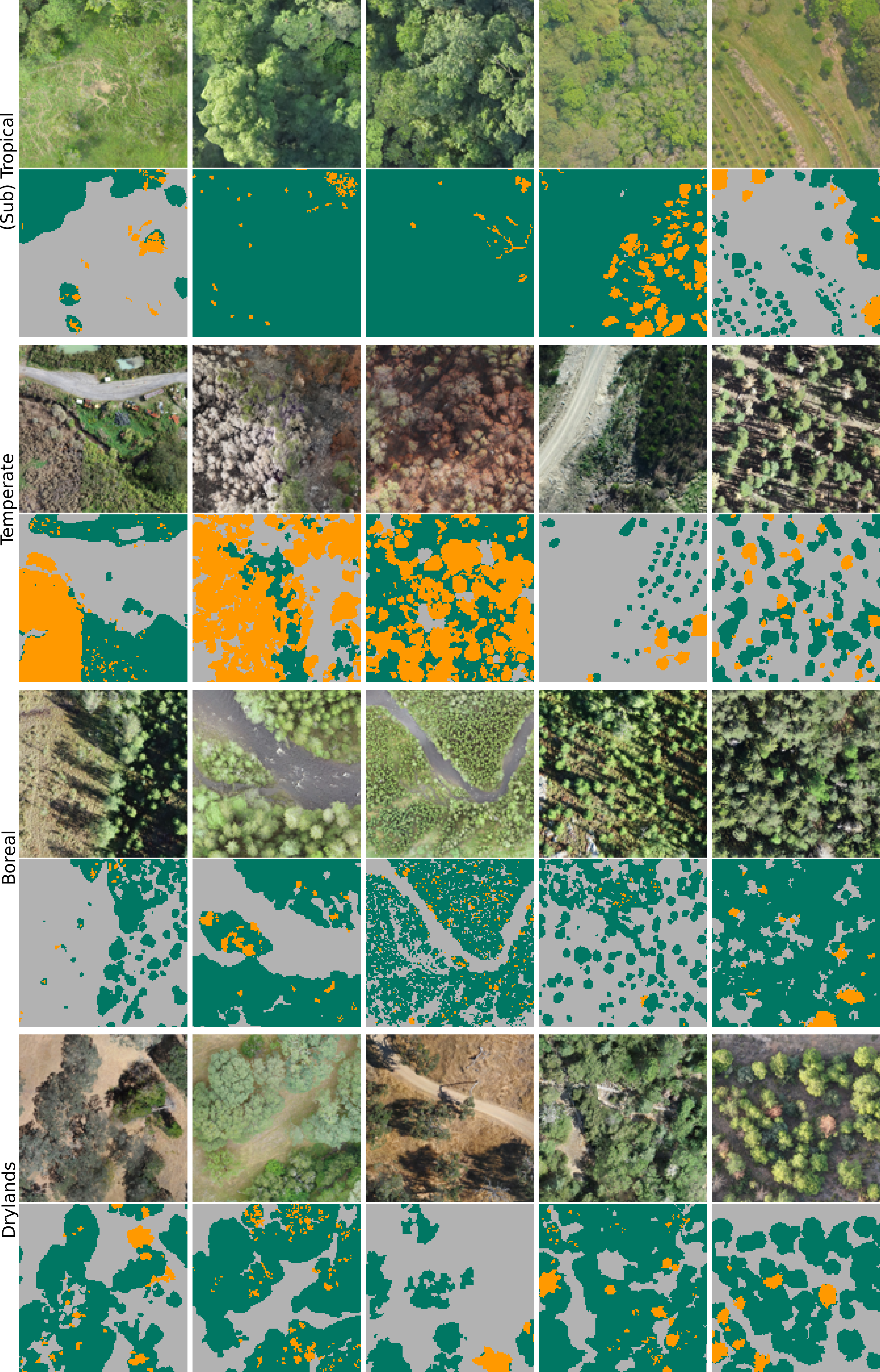}
    \caption{
Qualitative examples of aerial imagery tiles and corresponding semantic masks (tree cover: teal, mortality: orange, background: gray) across aggregated biome groups, illustrating diverse canopy structures across ecosystems.
    }
    \label{fig:visual_diversity_biome_and_fire}
\end{figure}

\section{Datasets: DTE-aerial-train and DTE-aerial-bench}

To facilitate large-scale forest monitoring, %
we curate DTE-aerial-train and DTE-aerial-bench, which are the first globally distributed training and benchmark datasets of centimeter-scale aerial imagery with joint annotations of tree cover and mortality (Table~\ref{tab:dataset_comparison}). 
These datasets are constructed from the deadtrees.earth platform \citep{mosig2026deadtrees}, where imagery is crowd-sourced globally through collaborations with researchers, ecologists, and environmental monitoring initiatives. While DTE-aerial shares its underlying imagery source with the earlier deadtrees.earth mortality dataset \citep{mohring2025global}, its contributions lie in (i) a new multi-class annotation schema that jointly labels tree cover and mortality, (ii) large-scale model-assisted pseudo-label generation, (iii) a scale-up from 434 to 2,176 training sites, and (iv) a newly curated, expert-annotated benchmark.
Each orthophoto is audited by the deadtrees.earth team and accompanied by detailed metadata, including geocoordinates, biome, acquisition date, contributors, spatial resolution, and licensing information \citep{mosig2026deadtrees}.
Our dataset comprises permissively licensed imagery with spatial resolution finer than $\approx 20$\,cm, acquired between 2012 and 2025, and restricted to the growing season to enable robust discrimination between seasonal leaf senescence and tree mortality.

\paragraph{Multi-resolution.} 

To enable generalization across sensors and scales, we construct both the training and test datasets as multi-resolution datasets. The native data is strongly imbalanced toward high-resolution drone imagery (Fig.~\ref{fig:dataset_overview}). Following \citet{mohring2025global}, this allows us to generate patches at multiple resolutions via iterative downsampling, producing patches at all target resolutions (2.5, 5, 10, 20\,cm) coarser than or equal to the native resolution. This enables coverage from fine to coarse scales while preserving spatial extent (see Appendix for details).

For the benchmark dataset, we restrict orthophotos to resolutions finer than 5\,cm, ensuring that each orthophoto can generate corresponding patches at 5, 10, and 20\,cm. This design enables consistent cross-scale evaluation of the same underlying scene, minimizing confounding effects from varying native resolutions and ensuring that performance differences can be attributed to scale.
In the training set, all orthophotos are tiled into $1024 \times 1024$ patches with a 512-pixel overlap, following \citet{mohring2025global}. %
We cap the number of patches per orthophoto at 1,600 to mitigate bias for overly large orthophotos. In the benchmark set, we instead use non-overlapping $1024 \times 1024$ patches.

\paragraph{Spatial and ecological coverage.}
DTE-aerial-train comprises 2,176 globally distributed aerial images, corresponding to 346K image patches
(Fig.~\ref{fig:dataset_overview}). To reveal ecological coverage, we assign each orthophoto to established global biome definitions based on the coordinates of its centroid \citep{olson2001terrestrial}. For analysis, biomes are aggregated into four broad groups: temperate, (sub)tropical, drylands (including Mediterranean), and boreal and montane, following \citet{mosig2026deadtrees}. The training dataset is dominated by  %
temperate regions with 1,206 orthophotos ($\sim$171K patches), followed by (sub)tropical regions  (722 orthophotos, $\sim$135K patches), drylands (194 orthophotos, $\sim$35K patches) and boreal and montane regions (54 orthophotos, $\sim$5K patches). %

DTE-aerial-bench comprises 25 orthophotos (525 patches) covering facets of forest structure (e.g., managed and unmanaged forests), biome groups (including 11 temperate, 4 (sub)tropical, 6 dryland, and 4 boreal and montane sites), and mortality stages, where 4 sites include early mortality stages characterized by brown foliage and fire-related mortality that is more challenging to detect. %
Overall, the dataset captures broad ecological variability across species composition, forest structure, and environmental gradients (Fig.~\ref{fig:visual_diversity_biome_and_fire}).
\begin{figure}[tp]
    \centering
    \includegraphics[width=\columnwidth]{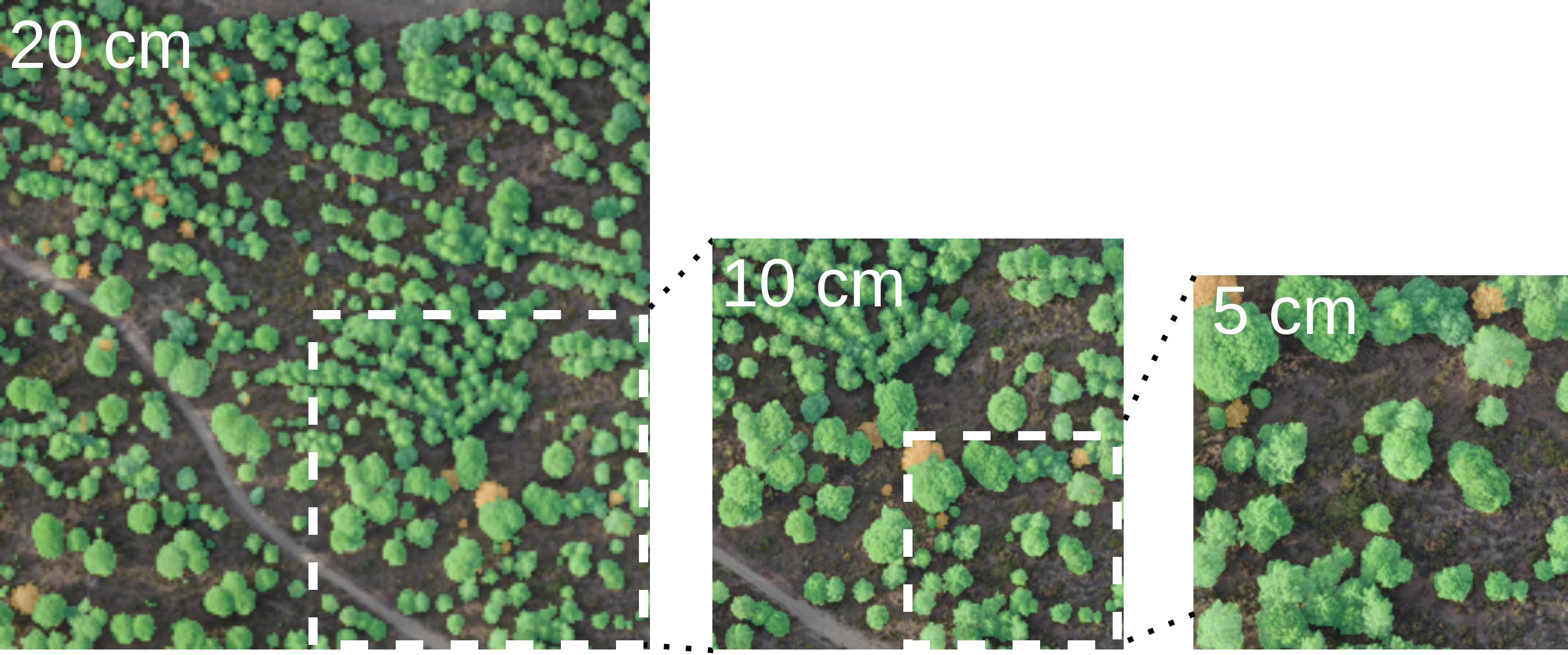}
    \caption{Multi-scale annotation and resampling in DTE-aerial-bench. Labels are created at 5\,cm spatial resolution (right) and propagated to 10\,cm and 20\,cm via downsampling. This ensures consistent annotations across spatial scales. We visualize
all patches of the benchmark dataset across resolutions on \href{https://deadtrees.earth/releases/dte-aerial-bench}{deadtrees.earth/releases/dte-aerial-bench}.
    }
    \label{fig:tiling}
\end{figure}

\paragraph{Annotation pipeline.}
We define three semantic annotation classes: (1) tree cover, (2) mortality, and (3) background. The tree cover class includes standing tree canopies of any size and condition. The mortality class captures a continuum of canopy decline in standing trees from partial dieback to fully dead tree crowns. %
Background includes scene components such as canopy gaps, bare ground, understory, and other non-tree areas. All classes are annotated via polygons, which enables precise delineation of these classes while remaining more scalable than dense pixel-wise labeling. %

The annotation effort required for this task is substantial. Incorporating partial crown dieback into the mortality class is essential for capturing early-stage tree mortality \citep{international2025towards, mohring2025global}, but significantly increases complexity, as partially affected crowns require fine-grained, fragmented, and irregular polygons compared to fully dead trees (e.g., Fig.~\ref{fig:visual_diversity_biome_and_fire}, (sub)tropical row, columns 2 and 3; dryland row, columns 2 and 4). Accurately delineating background regions further adds to this challenge. In particular, canopy gaps, defined as openings in the forest canopy and critical indicators of forest vitality and degradation \citep{hagemann2022analysing, perry2008forest}, require precise boundary annotation (e.g., Fig.~\ref{fig:visual_diversity_biome_and_fire}, boreal row, column 5).
Together, these factors make the labeling process considerably more demanding than conventional binary annotation.

To balance annotation quality with scalability, we adopt a two-stage annotation strategy. For the benchmark test set, we prioritize high annotation quality by curating an ecologically diverse set of aerial images that are exhaustively labeled across all classes. These images are repeatedly annotated by multiple domain experts, with disagreements resolved through iterative consensus-building to ensure high precision and consistency. All annotations are performed at the highest available resolution (5\,cm), and labels at coarser resolutions (e.g., 10\,cm, 20\,cm) are systematically derived from these base annotations, ensuring consistent cross-scale evaluation (Fig.~\ref{fig:tiling}). Depending on scene complexity, sites underwent up to five rounds of review and refinement before consensus was reached.

For the training data, we rely on coarsely annotated imagery to achieve extensive geographic coverage while maintaining tractable annotation costs. Labels are initialized for a common set of orthophotos using manually audited pseudo-labels derived with segmentation models of \citet{mohring2025global} for mortality and \citet{veitch2024oam} for tree cover.
As \citet{veitch2024oam} does not reliably capture canopy gaps, we reassign pixels initially labeled as tree cover to background where the vegetation signal is low using heuristic thresholds. The thresholds were selected empirically through iterative visual inspection by expert annotators across orthophotos spanning all biome groups. Specifically, pixels with grayscale intensity below 25 (0--255) were grouped into connected regions, and regions containing at least 50 pixels were reassigned to background.

Experts first performed an orthophoto-level audit of all initial pseudo-labels, and only orthophotos with satisfactory overall label quality were retained for training. The retained pseudo-labels were then iteratively improved through a model-assisted relabeling process: at each iteration, a model was trained on the current labels and applied to the dataset. For each orthophoto, auditors reviewed a randomly selected subset of 25 patches, comparing the existing labels with the updated model predictions. Reviewers searched in particular for recurring (systematic) error patterns, such as false positives on trees with colored phenology or seasonal discoloration, incomplete crown segmentation, and missing or partial segmentation of dieback. If the updated predictions consistently resolved previously observed errors, they were adopted as the new pseudo-labels for that orthophoto; otherwise, the orthophoto was re-examined after subsequent model iterations, with auditors selecting the most accurate predictions based on visual inspection. 
The intermediate models therefore served as an aid to accelerate label refinement rather than as a fully automated label-generation pipeline.
Additionally, to better capture underrepresented and visually distinct mortality stages, we include a set of eight expert-annotated sites covering early mortality stages characterized by reddish or brown foliage and fire-related mortality \citep{mosig2026deadtrees, mohring2025global}. 

In total, 4 annotators and 4 domain experts contributed to dataset curation. Annotators were trained by the domain experts to ensure consistent interpretation of tree mortality signals. Domain experts were closely involved throughout the process, including quality control, disagreement resolution, and iterative feedback. The overall labeling effort amounted to approximately 550 paid person-hours.

\paragraph{Data licensing and usage.}
To enable broad reuse and support open research, we design our dataset around permissive licensing principles. The dataset comprises openly licensed imagery, with most samples released under permissive licenses such as Creative Commons Attribution 4.0 (CC BY 4.0) and the MIT License. Specifically, among the 2,176 aerial images in DTE-aerial-train, 2,162 are licensed under CC BY 4.0, 9 under the MIT License, and 5 under Creative Commons Attribution--NonCommercial--ShareAlike (CC BY-NC-SA).
We distribute all image tiles under the same license as their corresponding source imagery.

\paragraph{Intended use.}
The dataset is designed for developing and \textit{rigorously evaluating} machine learning models for large-scale segmentation of tree cover and tree mortality. The expert-annotated benchmark test set enables standardized comparison across methods, while the larger, coarsely labeled training set reflects realistic constraints such as label noise, class imbalance, and incomplete annotations. Together, they provide a reproducible benchmark for assessing model robustness and generalization across geographic regions, spatial resolutions, and disturbance regimes.

\paragraph{ML problem and ML-ready dataset.}
Our datasets introduce several core challenges for the computer vision community, including (i) long-tailed spatial distributions, (ii) multi-resolution variability across image acquisition scales, (iii) noisy supervision from coarsely labeled data, and (iv) rare out-of-distribution disturbance scenarios. These properties are primarily captured in DTE-aerial-train, which reflects real-world data variability, while DTE-aerial-bench serves as a carefully curated testbed with high-quality expert annotations. Together, they define a realistic and controlled benchmark for studying robustness, generalization, and uncertainty-aware learning. In addition, the dataset enables new research directions such as cross-biome generalization under limited training data, transfer across spatial resolutions, and zero-shot adaptation of foundation models to high-resolution aerial imagery.

\section{Experiments}

\paragraph{Dataset pre-processing and augmentation.}

We randomly split orthophotos of DTE-aerial-train into 90\% training and 10\% validation, resulting in 1,959 orthophotos (311K patches) for training and 217 orthophotos (35K patches) for validation. None of the 25 DTE-aerial-bench orthophotos is contained in the training or validation split; the benchmark identifiers are listed in Appendix~\ref{sec:bench_ids}. All models are trained on random crops of size $640 \times 640$ pixels (ablations in Appendix~\ref{sec:ablation_patch_size}). For regularization, we apply a series of data augmentations, including horizontal and vertical flips, 90-degree rotations, grayscale conversion, gamma adjustments, and color jittering. Finally, input images are normalized using ImageNet statistics.

\paragraph{Backbones and architectures.}
We evaluate a diverse set of segmentation architectures spanning convolutional, transformer-based, and foundation model approaches. For convolutional baselines, we adopt U-Net~\citep{ronneberger2015u} and DeepLabV3+ (DLV3+)~\citep{chen2018encoder} with ResNet-34 and ResNet-50 encoders, implemented using the Segmentation Models PyTorch library~\citep{iakubovskii2019segmentation}. These models are initialized with ImageNet pre-trained weights, which improves convergence over training from scratch.
For transformer-based models, we include SegFormer~\citep{xie2021segformer} with MiT-B1, and MiT-B3 variants, capturing a range of model capacities. We further evaluate Mask2Former (M2F)~\citep{cheng2022masked} in its tiny and small configurations to assess performance under mask-based decoding frameworks. These models are initialized from weights pre-trained on ADE-20K.

Finally, we incorporate foundation model features using the DINOv2 base model with two adaptation strategies~\citep{oquab2023dinov2}. In the single-layer setting (DINOv2$^{\text{sl}}$), only the final transformer layer is used for decoding. In the multi-layer setting (DINOv2$^{\text{ml}}$), intermediate representations from layers $\{3, 6, 9, 12\}$ are aggregated to capture hierarchical features across the network. Each feature map is projected to 128 channels, concatenated, and passed to a lightweight decoder, enabling the combination of fine-grained spatial details with high-level semantic context in a computationally efficient manner.

\paragraph{Training.}

We use a Tversky loss~\citep{abraham2019novel} to address class imbalance. For the background and tree cover classes, we set equal weighting between precision and recall ($\alpha=0.5$, $\beta=0.5$). For the mortality class, which is rare and exhibits a long-tailed distribution, we emphasize recall by setting $\alpha=0.3$ and $\beta=0.7$, encouraging the model to better capture minority class instances.

\paragraph{Evaluation.}
Our evaluation framework is designed to reflect real-world deployment conditions by assessing model performance across (i) biome groups, (ii) resolution, and (iii) segmentation tasks. 
We report all results using the F1 score, and additionally IoU, precision, and recall. All metrics are computed independently for each site and then averaged across sites. Results are reported as the mean and standard deviation over three independent runs with different random seeds (0, 100, 200) to account for variability due to stochastic training dynamics and random initialization. For each metric, we aggregate performance across runs and present results in the form $\mu \pm 2\sigma$. Because seed variation only captures training stochasticity, we additionally assess site-level robustness with a leave-one-site-out analysis (Section~\ref{sec:loso}, Table~\ref{tab:loso}).

\paragraph{Environmental impact.}
Models were trained on a single NVIDIA H200 GPU, with the largest model using up to 120\,GB VRAM at a 600\,W power target and requiring approximately 25 hours of training. Using the Machine Learning Impact calculator~\citep{lacoste2019quantifying}, we estimate a total carbon footprint of $\sim$140\,kg CO$_2$ for all experiments and ablations, excluding testing and failed runs; this estimate likely underestimates total emissions.

\section{Results and discussion}
\begin{table}[tp]
\centering
\caption{
Comparison of models on mortality and tree cover segmentation performance on the benchmark dataset. We report aggregated F1 for biome groups and resolutions. Results are averaged over three seeds, with twice the standard deviation ($\pm 2\sigma$) shown in parentheses. Best results are highlighted in \textbf{bold} and second-best in \textit{italics}. DT-V1 is the model of
\citet{mohring2025global} and TCD that of \citet{veitch2024oam}.}
\setlength{\tabcolsep}{3pt}
\label{tab:mortalityandtreecover_biome_res_eval}
\fitwidth{%
\begin{tabular}{lccccccc}
\toprule
Model & Temperate & Tropical & Boreal & Drylands & 5\,cm & 10\,cm & 20\,cm \\
\midrule
\multicolumn{8}{c}{\textbf{Mortality (F1)}} \\
\midrule
DT-V1 & 0.51 & 0.64 & 0.40 & 0.55 & 0.55 & 0.47 & 0.39 \\

MiT-B3 & \textbf{0.58}\sd{0.04} & \textit{0.67}\sd{0.00} & \textbf{0.58}\sd{0.01} & 0.55\sd{0.03} & \textbf{0.60}\sd{0.02} & \textbf{0.56}\sd{0.03} & \textbf{0.44}\sd{0.02} \\
MiT-B1 & \textit{0.54}\sd{0.04} & \textit{0.67}\sd{0.01} & \textit{0.57}\sd{0.01} & \textit{0.58}\sd{0.00} & \textit{0.59}\sd{0.02} & 0.54\sd{0.03} & \textit{0.43}\sd{0.00} \\
U-Net{\scriptsize(res34)} & 0.52\sd{0.05} & 0.65\sd{0.01} & 0.53\sd{0.01} & \textbf{0.59}\sd{0.00} & 0.58\sd{0.02} & 0.53\sd{0.02} & 0.41\sd{0.02} \\
U-Net{\scriptsize(res50)} & 0.53\sd{0.03} & 0.64\sd{0.01} & 0.52\sd{0.03} & \textbf{0.59}\sd{0.02} & 0.58\sd{0.02} & 0.53\sd{0.01} & \textbf{0.44}\sd{0.02} \\
M2F{\scriptsize(s)} & 0.53\sd{0.08} & \textit{0.67}\sd{0.01} & 0.56\sd{0.02} & \textit{0.58}\sd{0.02} & 0.58\sd{0.03} & \textit{0.55}\sd{0.05} & \textit{0.43}\sd{0.01} \\
M2F{\scriptsize(t)} & \textit{0.54}\sd{0.05} & \textbf{0.68}\sd{0.00} & \textbf{0.58}\sd{0.01} & 0.57\sd{0.01} & \textit{0.59}\sd{0.02} & \textit{0.55}\sd{0.03} & \textbf{0.44}\sd{0.01} \\
DLV3+{\scriptsize(res34)} & 0.50\sd{0.08} & 0.66\sd{0.00} & 0.53\sd{0.02} & 0.56\sd{0.01} & 0.57\sd{0.04} & 0.49\sd{0.02} & 0.40\sd{0.02} \\
DLV3+{\scriptsize(res50)} & 0.52\sd{0.02} & 0.66\sd{0.01} & 0.56\sd{0.02} & 0.56\sd{0.01} & 0.58\sd{0.01} & 0.50\sd{0.02} & 0.41\sd{0.01} \\

DINOv2(b)$^{\text{ml}}$ & 0.41\sd{0.01} & 0.63\sd{0.01} & 0.41\sd{0.02} & 0.54\sd{0.01} & 0.48\sd{0.01} & 0.46\sd{0.01} & 0.35\sd{0.01} \\
DINOv2(b)$^{\text{sl}}$ & 0.39\sd{0.02} & 0.63\sd{0.01} & 0.40\sd{0.01} & 0.51\sd{0.01} & 0.47\sd{0.00} & 0.43\sd{0.01} & 0.35\sd{0.01} \\

\midrule

\multicolumn{8}{c}{\textbf{Tree cover (F1)}} \\
\midrule
TCD & \textbf{0.89} & \textit{0.91} & 0.91 & \textit{0.91} & \textbf{0.90} & \textbf{0.91} & \textit{0.87} \\

MiT-B3 & \textit{0.85}\sd{0.00} & \textbf{0.92}\sd{0.02} & \textbf{0.93}\sd{0.00} & \textbf{0.92}\sd{0.00} & \textit{0.89}\sd{0.00} & \textit{0.90}\sd{0.00} & \textbf{0.89}\sd{0.00} \\
MiT-B1 & \textit{0.85}\sd{0.00} & \textit{0.91}\sd{0.01} & \textbf{0.93}\sd{0.00} & \textbf{0.92}\sd{0.00} & \textit{0.89}\sd{0.00} & 0.89\sd{0.00} & \textbf{0.89}\sd{0.00} \\
U-Net{\scriptsize(res50)} & \textit{0.85}\sd{0.00} & \textbf{0.92}\sd{0.01} & \textit{0.92}\sd{0.00} & \textbf{0.92}\sd{0.00} & \textit{0.89}\sd{0.00} & \textit{0.90}\sd{0.00} & \textbf{0.89}\sd{0.00} \\
U-Net{\scriptsize(res34)} & \textit{0.85}\sd{0.01} & \textbf{0.92}\sd{0.00} & \textit{0.92}\sd{0.00} & \textbf{0.92}\sd{0.00} & \textit{0.89}\sd{0.01} & \textit{0.90}\sd{0.00} & \textbf{0.89}\sd{0.00} \\
M2F{\scriptsize(s)} & \textit{0.85}\sd{0.01} & \textit{0.91}\sd{0.03} & \textbf{0.93}\sd{0.00} & \textbf{0.92}\sd{0.01} & \textit{0.89}\sd{0.01} & \textit{0.90}\sd{0.01} & \textbf{0.89}\sd{0.01} \\
M2F{\scriptsize(t)} & 0.84\sd{0.01} & \textbf{0.92}\sd{0.01} & \textbf{0.93}\sd{0.00} & \textbf{0.92}\sd{0.01} & \textit{0.89}\sd{0.01} & 0.89\sd{0.00} & \textbf{0.89}\sd{0.01} \\
DLV3+{\scriptsize(res50)} & 0.84\sd{0.01} & \textit{0.91}\sd{0.00} & \textbf{0.93}\sd{0.00} & \textbf{0.92}\sd{0.00} & 0.88\sd{0.00} & \textit{0.90}\sd{0.01} & \textbf{0.89}\sd{0.00} \\
DLV3+{\scriptsize(res34)} & 0.84\sd{0.01} & \textit{0.91}\sd{0.01} & \textbf{0.93}\sd{0.00} & \textbf{0.92}\sd{0.00} & 0.88\sd{0.00} & 0.89\sd{0.00} & \textbf{0.89}\sd{0.00} \\
DINOv2(b)$^{\text{ml}}$ & \textit{0.85}\sd{0.00} & \textit{0.91}\sd{0.01} & 0.87\sd{0.01} & 0.88\sd{0.00} & 0.87\sd{0.00} & 0.87\sd{0.00} & 0.84\sd{0.01} \\
DINOv2(b)$^{\text{sl}}$ & 0.84\sd{0.01} & \textit{0.91}\sd{0.00} & 0.87\sd{0.01} & 0.88\sd{0.00} & 0.87\sd{0.00} & 0.86\sd{0.00} & 0.84\sd{0.01} \\

\bottomrule
\end{tabular}%
}

\end{table}

In Table \ref{tab:mortalityandtreecover_biome_res_eval}, we evaluate baseline models on DTE-aerial-bench across biomes and spatial resolutions to assess generalization under ecological and sensor variability. For mortality, compared to the DT-V1 \citep{mohring2025global}, most models trained on DTE-aerial-train improve mortality segmentation, although the size of the gains differs substantially between biomes. The strongest gains are observed in the boreal and montane biome group, with MiT-B3 achieving a +45\% relative improvement in boreal forests (0.40 $\rightarrow$ 0.58). Additional improvements are observed in temperate forests (+13.7\%, 0.51 $\rightarrow$ 0.58), while the gain in tropical forests (+4.7\%) is modest and MiT-B3 performs on par with the baseline in drylands (0.55). 

 Performance improvements are further consistent across spatial resolutions. Compared to DT-V1, MiT-B3 achieves a +9.1\% relative improvement at 5\,cm (0.55 $\rightarrow$ 0.60), +19.1\% at 10\,cm (0.47 $\rightarrow$ 0.56), and +12.8\% at 20\,cm (0.39 $\rightarrow$ 0.44). This trend suggests that our approach is particularly effective at leveraging contextual information when fine-grained spatial details are degraded. 

In addition to improved mortality detection, the joint setting enables consistent estimation of tree cover, allowing relative mortality to be quantified within forested areas.
For tree cover, F1 remains close to the OAM baseline \citep{veitch2024oam} across most biomes and resolutions, with a notable decrease in temperate forests (0.89 $\rightarrow$ 0.85).

Overall, MiT-B3 provides the best balance across both tasks, delivering substantial improvements for mortality (+11.3\%) at the cost of slightly lower tree-cover recall (Table~\ref{tab:class_level_eval}).

DINOv2-based models perform consistently worse than the other architectures, for both mortality and tree cover (Tables~\ref{tab:mortalityandtreecover_biome_res_eval} and~\ref{tab:class_level_eval}). As the DINOv2 backbone is pretrained on generic natural imagery and is not adapted to the aerial domain, we hypothesize that its representations transfer less effectively to centimeter-scale aerial imagery than those of models trained end-to-end on DTE-aerial-train, and the gap is largest for the fine-grained mortality class. 

\paragraph{Class-level performance.} Table~\ref{tab:class_level_eval} reports class-level performance using F1 score, Intersection over Union (IoU), precision, and recall. All models, except the foundation model (DINOv2), outperform the baseline on mortality detection. In particular, MiT-B3 achieves a substantial improvement for mortality (+11.3\% relative improvement), driven mainly by higher recall (0.47 $\rightarrow$ 0.54) at similar precision. For tree cover, however, all models trained on DTE-aerial-train have lower recall (0.92--0.93 vs.\ 0.96) and IoU (0.78--0.83 vs.\ 0.84) than OAM-TCD, while precision is similar or slightly higher.

\begin{table}[tp]
\centering
\small
\setlength{\tabcolsep}{2.1pt}
\caption{Class-level evaluation on mortality and tree cover. We
report F1, IoU, Precision and Recall for tree cover and mortality. DT-V1 is the model of
\citet{mohring2025global} and TCD that of \citet{veitch2024oam}.}
\label{tab:class_level_eval}

\fitwidth{%
\begin{tabular}{lcccccccc}
\toprule
& \multicolumn{4}{c}{\textbf{Mortality}} & \multicolumn{4}{c}{\textbf{Tree Cover}} \\
\cmidrule(lr){2-5} \cmidrule(lr){6-9}
Model & F1 & IoU & Prec. & Recall & F1 & IoU & Prec. & Recall \\
\midrule
DT-V1 & 0.53 & 0.39 & 0.69 & 0.47 & -- & -- & -- & -- \\
TCD & -- & -- & -- & -- & \textbf{0.90} & \textbf{0.84} & 0.86 & \textbf{0.96} \\

\midrule

MiT-B3 & \textbf{0.59}\sd{0.02} & \textbf{0.45}\sd{0.02} & \textit{0.71}\sd{0.02} & \textbf{0.54}\sd{0.03} & \textit{0.89}\sd{0.00} & \textit{0.83}\sd{0.00} & \textbf{0.88}\sd{0.01} & 0.92\sd{0.01} \\

MiT-B1 & \textit{0.58}\sd{0.02} & \textit{0.44}\sd{0.02} & \textbf{0.72}\sd{0.01} & \textit{0.53}\sd{0.02} & \textit{0.89}\sd{0.00} & 0.82\sd{0.00} & \textit{0.87}\sd{0.01} & \textit{0.93}\sd{0.01} \\
U-Net{\scriptsize(res50)} & 0.56\sd{0.01} & 0.42\sd{0.01} & 0.69\sd{0.02} & \textit{0.53}\sd{0.03} & \textit{0.89}\sd{0.00} & 0.82\sd{0.00} & \textit{0.87}\sd{0.00} & 0.92\sd{0.00} \\
U-Net{\scriptsize(res34)} & 0.56\sd{0.02} & 0.42\sd{0.02} & 0.69\sd{0.00} & 0.52\sd{0.02} & \textit{0.89}\sd{0.00} & 0.82\sd{0.00} & \textit{0.87}\sd{0.01} & \textit{0.93}\sd{0.01} \\

M2F{\scriptsize(s)} & 0.57\sd{0.04} & 0.43\sd{0.03} & \textit{0.71}\sd{0.03} & \textit{0.53}\sd{0.05} & \textit{0.89}\sd{0.01} & 0.82\sd{0.01} & \textit{0.87}\sd{0.02} & 0.92\sd{0.03} \\
M2F{\scriptsize(t)} & \textit{0.58}\sd{0.02} & \textit{0.44}\sd{0.02} & \textit{0.71}\sd{0.01} & \textit{0.53}\sd{0.02} & \textit{0.89}\sd{0.00} & 0.82\sd{0.01} & \textbf{0.88}\sd{0.01} & 0.92\sd{0.02} \\
DLV3+{\scriptsize(res50)} & 0.56\sd{0.01} & 0.42\sd{0.01} & 0.70\sd{0.01} & 0.51\sd{0.01} & \textit{0.89}\sd{0.00} & 0.82\sd{0.00} & 0.86\sd{0.01} & \textit{0.93}\sd{0.01} \\
DLV3+{\scriptsize(res34)} & 0.55\sd{0.03} & 0.41\sd{0.03} & 0.70\sd{0.01} & 0.50\sd{0.04} & 0.88\sd{0.00} & 0.81\sd{0.01} & 0.86\sd{0.01} & \textit{0.93}\sd{0.01} \\
DINOv2(b)$^{\text{ml}}$ & 0.47\sd{0.01} & 0.34\sd{0.00} & 0.64\sd{0.02} & 0.43\sd{0.02} & 0.87\sd{0.00} & 0.78\sd{0.00} & 0.83\sd{0.01} & 0.92\sd{0.01} \\
DINOv2(b)$^{\text{sl}}$ & 0.46\sd{0.00} & 0.33\sd{0.00} & 0.63\sd{0.02} & 0.43\sd{0.01} & 0.86\sd{0.00} & 0.78\sd{0.00} & 0.83\sd{0.01} & 0.92\sd{0.01} \\

\bottomrule
\end{tabular}%
}
\end{table}

\begin{figure}[tp]
    \centering
    \includegraphics[width=\columnwidth]{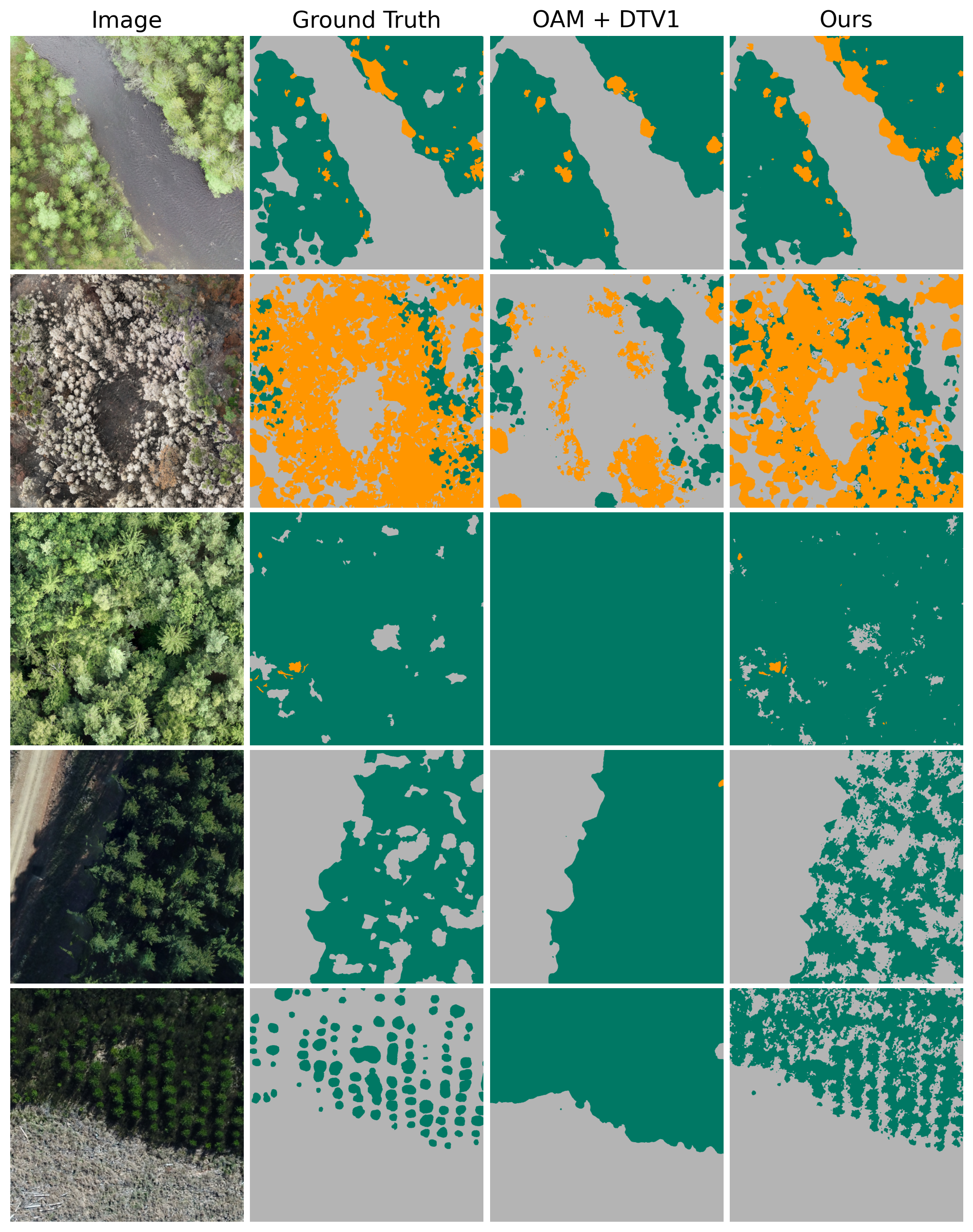}
    \caption{Qualitative comparison of segmentations on benchmark patches. %
    Our method produces more consistent delineation of tree cover and mortality than combined baselines from OAM and DTV1 \citep{veitch2024oam, mohring2025global}. 
    }

    \label{fig:qualitative}
\end{figure}

\paragraph{Tree-cover trade-off.} To understand the source of the lower tree-cover recall, we compare normalized confusion matrices of MiT-B3 and the combined baseline (Table~\ref{tab:confusion}). For the baseline, we construct a three-class prediction by first assigning tree cover from OAM-TCD \citep{veitch2024oam} and then overwriting pixels predicted as mortality by DT-V1 \citep{mohring2025global}; all remaining pixels are background. 
For our model, tree-cover errors occur predominantly as tree cover $\rightarrow$ background (5.0\%) and rarely for tree cover $\rightarrow$ mortality (0.8\%).
Manual inspection of the patches with the largest tree cover $\rightarrow$ background differences shows that our model segments considerably more canopy gaps and tighter canopy boundaries, particularly at forest edges, in plantation rows, and in fragmented canopies (Fig.~\ref{fig:qualitative}, last two rows). The baseline instead tends to predict large contiguous tree-cover regions that ignore small clearings. This raises its tree-cover recall (98.0\% vs.\ 94.2\%) but also assigns substantially more background to tree cover (17.0\% vs.\ 9.9\%). Users should therefore expect models trained on DTE-aerial-train to delineate gaps and edges more finely, at the cost of slightly lower recall at the perimeter of large continous canopies. 

\begin{table}[tp]
\centering
\caption{Normalized confusion matrices on DTE-aerial-bench (rows: ground truth; columns: prediction). The baseline combines OAM-TCD (tree cover) and DT-V1 (mortality).}
\label{tab:confusion}
\small
\begin{tabular}{lccc@{\hspace{2em}}ccc}
\toprule
& \multicolumn{3}{c}{\textbf{MiT-B3 (ours)}} & \multicolumn{3}{c}{\textbf{OAM-TCD + DT-V1}} \\
\cmidrule(lr){2-4} \cmidrule(lr){5-7}
Ground truth & Backgr. & Tree cover & Mortality & Backgr. & Tree cover & Mortality \\
\midrule
Background & 0.890 & 0.099 & 0.011 & 0.822 & 0.170 & 0.007 \\
Tree cover & 0.050 & 0.942 & 0.008 & 0.011 & 0.980 & 0.009 \\
Mortality  & 0.094 & 0.245 & 0.661 & 0.136 & 0.360 & 0.503 \\
\bottomrule
\end{tabular}
\end{table}

\paragraph{Site-level mortality fraction.} Pixel-level metrics do not directly show whether a model supports ecologically meaningful site-level quantities. We therefore compute, for each benchmark site, the mortality fraction $f = A_{\text{mort}} / (A_{\text{mort}} + A_{\text{tree}})$, i.e., the mortality area relative to the total area of tree cover and mortality. Each site is covered by 16 tiles at 5\,cm, 4 tiles at 10\,cm, and 1 tile at 20\,cm, all spanning the same geographic area, so $f$ is computed over identical extents at each resolution. We compare predicted and ground-truth fractions across the 25 sites using the coefficient of determination ($R^2$) and mean absolute error (MAE). MiT-B3 achieves high agreement at all resolutions (Table~\ref{tab:site_fraction}). 

\begin{table}[tp]
  \centering
  \caption{Agreement between predicted and reference site-level mortality
  fractions $f = A_\mathrm{mort}/(A_\mathrm{mort}+A_\mathrm{tree})$ on
  DTE-aerial-bench ($n = 25$ sites). For the baseline, mortality is taken from
  DT-V1 and tree cover from OAM-TCD; pixels predicted as deadwood by DT-V1 are counted as
  mortality only, not as tree cover.}
  \label{tab:site_fraction}
  \begin{tabular}{llccc}
    \toprule
     & Model & 5\,cm & 10\,cm & 20\,cm \\
    \midrule
    {$R^2$ $\uparrow$}
     & DT-V1 + OAM-TCD & 0.692 & 0.416 & 0.022 \\
     & MiT-B3  (Ours)        & \textbf{0.928} & \textbf{0.942} & \textbf{0.880} \\
    \midrule
    {MAE $\downarrow$}
     & DT-V1 + OAM-TCD & 0.036 & 0.047 & 0.065 \\
     & MiT-B3  (Ours)        & \textbf{0.022} & \textbf{0.020} & \textbf{0.033} \\
    \bottomrule
  \end{tabular}
\end{table}
\begin{table}[t]
\centering
\footnotesize
\setlength{\tabcolsep}{4pt}
\caption{\textbf{Sensitivity of DTE-aerial-bench to site selection} based on leave-one-site-out evaluation with MiT-B3. \textbf{F1} and \textbf{IoU} report the ranges of macro-averaged scores across runs, each excluding one orthophoto from the respective set. $\boldsymbol{\Delta}$\textbf{F1} reports the range of differences in F1 between MiT-B3 and the OAM-TCD + DT-V1 baseline, with the same orthophoto excluded in each comparison; positive values indicate higher F1 for MiT-B3.
}
\label{tab:loso}
\begin{tabular}{lc ccc ccc}
\toprule
 & & \multicolumn{3}{c}{Mortality} & \multicolumn{3}{c}{Tree cover} \\
\cmidrule(lr){3-5}\cmidrule(lr){6-8}
 & Sites & F1 & IoU & $\Delta$F1 & F1 & IoU & $\Delta$F1 \\
\midrule
All sites & 25 & 0.59--0.62 & 0.44--0.48 & $+$0.05 to $+$0.08 & 0.89--0.92 & 0.82--0.85 & 0.00 to $+$0.02 \\
\midrule
Temperate & 11 & 0.57--0.66 & 0.44--0.52 & $+$0.03 to $+$0.11 & 0.84--0.91 & 0.76--0.84 & $-$0.02 to $+$0.01 \\
Tropical  & 4 & 0.63--0.73 & 0.47--0.57 & $+$0.01 to $+$0.05 & 0.90--0.97 & 0.82--0.93 & 0.00 to $+$0.03 \\
Drylands  & 6 & 0.51--0.61 & 0.36--0.46 & $-$0.03 to $+$0.04 & 0.92--0.93 & 0.85--0.88 & $+$0.01 to $+$0.02 \\
Boreal    & 4 & 0.53--0.64 & 0.37--0.48 & $+$0.11 to $+$0.23 & 0.92--0.94 & 0.85--0.89 & $+$0.02 to $+$0.03 \\
\bottomrule
\end{tabular}
\end{table}
\paragraph{Sensitivity to site selection.}
\label{sec:loso}
To assess whether the benchmark results depend on individual sites, we performed
a leave-one-site-out analysis. For the full benchmark, all metrics were computed
25 times, each time on the 24 orthophotos remaining after excluding one; within
each biome group of $n$ orthophotos, they were computed $n$ times on the remaining
$n-1$ (Table~\ref{tab:loso}). The benchmark yields stable
scores and a consistent model ranking. Across the 25 runs on the full benchmark,
the mortality F1 of MiT-B3 ranged from 0.59 to 0.62 and its tree-cover F1 from
0.89 to 0.92, and MiT-B3 exceeded the OAM-TCD + DT-V1 baseline in mortality F1 in
every run (by 0.05 to 0.08). The exclusion of any single orthophoto therefore did
not alter the conclusions drawn from the benchmark.

Sensitivity to site selection differed among biomes. For tree mortality, F1
varied most in boreal forests (0.53--0.64) and least in temperate forests
(0.57--0.66), the group with the largest number of orthophotos. The improvement
over the baseline was greatest in boreal forests ($\Delta$F1 $+$0.11 to $+$0.23),
where the baseline performed weakest, and it persisted regardless of which
orthophoto was excluded.  In temperate forests, the
improvement was smaller but likewise remained positive in all runs ($+$0.03 to
$+$0.11). In tropical forests ($+$0.01 to $+$0.05), the improvement was small but positive in all runs; in drylands
($-$0.03 to $+$0.04), it was small and fell below zero
when individual orthophotos were excluded. Tree-cover F1 was stable within the
dryland and boreal groups (0.92--0.93 and 0.92--0.94) but more variable within the
temperate and tropical groups (0.84--0.91 and 0.90--0.97), with the lowest values
in temperate forests, consistent with the tree-cover results in
Table~\ref{tab:mortalityandtreecover_biome_res_eval}. As the biome groups comprise only four to eleven
orthophotos, these group-level comparisons should be regarded as indicative.
Overall, the advantage of MiT-B3 in detecting tree mortality was robust to the
exclusion of individual sites, whereas for tree cover the two approaches
performed comparably.

\section{Limitations}
\label{sec:limitations}
Several limitations should be considered when using the datasets. First, and most fundamentally, the input modality is RGB imagery, so the mortality labels describe visible mortality symptoms and crown decline as interpreted by experts, not physiological tree death. These cues can be confounded with phenology (e.g., seasonal leaf discolouration), illumination, shadow, stress-induced defoliation, and species-specific crown colour.
Therefore, phenological information of the respective site and local expertise may be required to accurately interpret the mortality delineations.
Second, the training labels are audited pseudo-labels derived from existing models; as no independent expert labels exist for the training split (apart from eight rare-mortality sites), residual label noise and biases inherited from the initial models cannot be quantified directly. The improvements on the independent expert benchmark indicate that the refined labels provide better supervision than the initial models, but do not rule out such biases. Third, both datasets are unevenly distributed across biomes and regions because they depend on crowd-sourced data availability; the benchmark contains 25 independent sites (11 temperate, 4 (sub)tropical, 6 dryland, 4 boreal and montane), so per-biome results, in particular for boreal and montane forests, are subject to site-level uncertainty.
Finally, the training dataset is biased toward high-resolution imagery (<10 cm), making models less directly transferable to lower-resolution (>30 cm) airplane or satellite data without adaptation

\section{Conclusion}

We introduced DTE-aerial-train and DTE-aerial-bench, a globally distributed, multi-resolution aerial image resource for joint segmentation of tree cover and mortality. By combining large-scale, coarsely annotated training data with an expert-annotated benchmark test set, the dataset enables standardized evaluation under realistic variation in biome, spatial resolution, and mortality expression. Our baselines demonstrate that training on DTE-aerial-train improves mortality detection, while tree-cover F1 remains close to a strong baseline with a trade-off in recall. The new model substantially improved estimation of the proportion of total tree crown area affected by mortality, with mean \(R^2\) increasing by 143\% (from 0.377 to 0.917) and mean absolute error decreasing by 49\% (from 0.049 to 0.025) across the three spatial resolutions. We hope this resource supports reproducible progress in robust forest monitoring and serves as a challenging benchmark for advancing generalization, uncertainty-aware learning, and segmentation under real-world distribution shifts.

\impact{We expect the release of these datasets to support large-scale ecological monitoring of tree cover and mortality, including research on forest responses to drought, pests, and fire, as well as the development of models that can inform forest management and Earth system modeling. Potential negative consequences primarily relate to the incidental capture or misuse of imagery. Aerial imagery may contain private land, buildings, or other infrastructure. Imagery is contributed to the deadtrees.earth platform by researchers and monitoring initiatives, whose Terms of Service require contributors to hold all necessary rights, avoid personal data of third parties, including identifiable individuals, and comply with applicable data-protection law. Acquisitions target forested ecosystems, and all annotation classes relate solely to vegetation; man-made structures are therefore incidental to the intended data collection. We do not systematically screen imagery for such content and cannot rule out that individual images contain buildings or other structures. The imagery and derived models should therefore not be repurposed for surveillance or property-related applications.}

\acks{C.M., T.K., and J.V.J. thank the European Space Agency for funding the
``DeepFeatures'' project via the AI4SCIENCE activity and the ``FORTRACK''
project via the ESA CLIMATE SPACE: Climate--Biodiversity studies. C.M.
acknowledges financial support from the German Aerospace Centre (DLR) on behalf
of the Federal Ministry for Economic Affairs and Climate Action (BMWK) for
project ML4Earth (FKZ 50EE2201B). T.K. acknowledges funding from the German
Research Foundation via the ``PANOPS'' project (grant no. 504978936) and
EXC 3127 Future Forests. T.K. and J.V.J. acknowledge support by the Ministry for
Food, Rural Areas, and Consumer Protection, Baden-W\"urttemberg, Germany (grant
no. 52-8670.00) and the German Research Foundation via the NFDI4Earth pilot
project GeoLabel (DFG grant no. 460036893).  We thank all
contributors who shared their orthophotos via deadtrees.earth.
The authors declare no competing interests.
}
\vskip 0.2in

\bibliography{main}

\appendix
\section{Data appendices}

\subsection{Data preprocessing - DTE-aerial-train}

We applied a standardized preprocessing pipeline consisting of polygon label rasterization, reprojection, multi-scale resampling, and tiling to transform heterogeneous raw data from \textit{deadtrees.earth} into a consistent format suitable for model training.

\paragraph{Label rasterization.}
Following \citet{mohring2025global}, vector annotations delineating mortality and tree cover are converted into pixel-wise masks aligned with the original orthophotos. Pixels corresponding to tree cover are assigned a value of 1, mortality pixels a value of 2, while background pixels are assigned 0.

The datasets obtained from \textit{deadtrees.earth} provide careful labels of high- and low-quality image regions, labeled as inside and outside of the Area of Interest (AOI) during the platform's data audit \citep{mosig2026deadtrees}. Here, regions outside the AOI are marked with a void label (255) and excluded from both loss computation and evaluation. This allows the use of tiles that partially extend beyond annotated regions without introducing training noise.
Inclusion of such partial tiles is critical to creating a robust prediction model that can handle the often jagged border of drone orthophotos.

\paragraph{Reprojection.}
To harmonize spatial representations, we reproject all orthophotos and corresponding masks to their respective Universal Transverse Mercator (UTM) zones, determined from the WGS84 centroid of each image \citep{mohring2025global}. The resampling was performed using GDAL with cubic convolution resampling and nearest neighbor resampling for imagery and masks, respectively. This transformation places all data in a consistent metric and local coordinate system (meters), ensuring that spatial patterns, object sizes, and distances are represented uniformly across samples. Such consistency is critical for learning robust spatial features, as it prevents scale ambiguities and geometric distortions that could otherwise degrade model training and limit generalization across sites and sensors \citep{schiefer2023uav}.%

\paragraph{Multi-scale resampling.}
The crowd-sourced drone and airplane imagery in \textit{deadtrees.earth} spans multiple spatial resolutions due to varying flight altitudes and sensors. To improve robustness to this variability, we augment the dataset through multi-scale resampling \citep{mohring2025global}. Each orthophoto is resampled to a fixed set of target spatial resolutions, 2.5\,cm, 5\,cm, 10\,cm, and 20\,cm. For instance, an image originally captured at 1.5\,cm~is downsampled to all coarser resolutions, exposing the model to a broad range of spatial scales during training.

\paragraph{Tiling.}
All images and masks are partitioned into fixed-size tiles of $1024 \times 1024$ pixels, extracted with 50\% overlap. Tiles containing more than 30\% void pixels are discarded. We retain tiles with no mortality instances, as they provide essential negative examples that help the model generalize to non-forest regions and healthy canopy areas. After tiling, we sample at most 1,600 tiles per orthophoto to mitigate bias from overly large images. Sampling is balanced across resolutions, with 400 patches drawn from each spatial resolution (2.5\,cm, 5\,cm, 10\,cm, and 20\,cm). Within each resolution, we further enforce class balance by selecting 200 patches containing mortality instances (>0 pixels) and 200 without.

\subsection{Data preprocessing - DTE-aerial-bench}
For DTE-aerial-bench, we developed a web-based annotation tool on the \textit{deadtrees.earth} platform to effectively assist annotators in labeling the large data files, while also enabling collaborative and iterative labeling. In this web-interface, annotators labeled high-resolution aerial imagery at 5\,cm spatial resolution using a fixed set of 16 tiles per site. Labels at coarser resolutions (10\,cm and 20\,cm) were generated automatically through downsampling (see Fig.~\ref{fig:tiling}), resulting in 16 tiles at 5\,cm, 4 tiles at 10\,cm, and 1 tile at 20\,cm per site, and a total of 21 tiles per site. The annotation tool also incorporated a quality control workflow with multi-reviewer validation. Only patches approved by multiple reviewers for both mortality and tree cover were retained in the final benchmark.

\subsection{Benchmark site identifiers}
\label{sec:bench_ids}
DTE-aerial-bench consists of the following 25 deadtrees.earth orthophoto identifiers: 375, 400, 435, 868, 1371, 1381, 1396, 1406, 3251, 3341, 3834, 4010, 4087, 4088, 4181, 4182, 4471, 5387, 5463, 5584, 5756, 5783, 5786, 5931, and 6445.

\section{Technical appendices and supplementary material}
\subsection{Hyperparameters and early stopping}
We train all models with a batch size of 32 using the AdamW optimizer with weight decay 0.01 and momentum parameters $\beta_1 = 0.9$ and $\beta_2 = 0.99$. A cosine learning rate schedule is used, decaying from $1 \times 10^{-4}$ to $1 \times 10^{-5}$ over 100K training iterations.

We use early stopping based on the mean per-site F1-score, computed by averaging F1-scores across sites to reduce bias toward sites with many patches. Training stops if either the mortality F1-score or the mean F1-score across all classes fails to improve by at least $1 \times 10^{-3}$ over two consecutive evaluation epochs (approximately 19K iterations).

\subsection{Ablation on loss function}
In Tables~\ref{tab:ablation_patch} and~\ref{tab:loss_pr}, we analyze the impact of the loss function. Given the long-tailed and spatially heterogeneous nature of mortality detection, we adopt Tversky loss to bias optimization toward recall ($\alpha=0.3$, $\beta=0.7$), improving detection of the rare mortality class while tolerating controlled false positives. We compare this to Dice loss, which equally weights precision and recall (i.e., $\alpha=\beta=0.5$). Tversky loss improves mortality detection across all spatial scales, with modest gains across most biomes and the largest improvement in temperate regions (+23.4\% relative gain, 0.47 $\rightarrow$ 0.58). 

\begin{table}[tp]
\centering
\caption{Class-level metrics of MiT-B3 trained with Tversky and Dice loss ($\mu \pm 2\sigma$ over three seeds).}
\label{tab:loss_pr}
\fitwidth{%
\begin{tabular}{lcccccccc}
\toprule
& \multicolumn{4}{c}{\textbf{Mortality}} & \multicolumn{4}{c}{\textbf{Tree cover}} \\
\cmidrule(lr){2-5} \cmidrule(lr){6-9}
Loss & F1 & IoU & Prec. & Recall & F1 & IoU & Prec. & Recall \\
\midrule
Tversky & 0.59\sd{0.02} & 0.45\sd{0.02} & 0.71\sd{0.02} & 0.54\sd{0.03} & 0.89\sd{0.00} & 0.83\sd{0.00} & 0.88\sd{0.01} & 0.92\sd{0.01} \\
Dice & 0.53\sd{0.02} & 0.40\sd{0.01} & 0.76\sd{0.00} & 0.45\sd{0.02} & 0.89\sd{0.00} & 0.82\sd{0.00} & 0.87\sd{0.00} & 0.92\sd{0.00} \\
\bottomrule
\end{tabular}%
}
\end{table}

\subsection{Ablation on image patch size}
\label{sec:ablation_patch_size}
To assess the impact of patch size, which determines the spatial context available to the model for capturing mortality and tree cover patterns, we perform an ablation over patch sizes of 224, 512, and 640 (Table~\ref{tab:ablation_patch}). For mortality, the three patch sizes perform within 0.01 F1 of each other at every spatial resolution, while performance across biomes remains comparable, with no single configuration consistently dominating. This suggests that, within the explored range, model performance is only weakly sensitive to the spatial context defined by patch size. For tree cover, performance remains largely stable across patch sizes, with only minor variations, indicating robustness to the choice of spatial context.

\begin{table}[tp]
\centering
\caption{
Ablation study on training patch size and loss function. We report aggregated F1 for biome groups and resolutions. Results are averaged over three seeds, with twice the standard deviation ($\pm 2\sigma$) shown in parentheses. Best results are highlighted in \textbf{bold}.
}
\setlength{\tabcolsep}{1pt}
\label{tab:ablation_patch}
\fitwidth{%
\begin{tabular}{ccccccccc}

\toprule
Model & Patch size & Temperate & Tropical & Boreal & Drylands & 5\,cm & 10\,cm & 20\,cm \\

\midrule
\multicolumn{9}{c}{\textbf{Mortality (F1)}} \\
\midrule
MiT-B3 {\small(dice)} & 640 & 0.47\sd{0.05} & 0.65\sd{0.01} & 0.57\sd{0.01} & 0.53\sd{0.02} & 0.54\sd{0.02} & 0.50\sd{0.01} & 0.41\sd{0.01} \\

MiT-B3 & 640 & \textbf{0.58}\sd{0.04} & \textbf{0.67}\sd{0.00} & 0.58\sd{0.01} & 0.55\sd{0.03} & \textbf{0.60}\sd{0.02} & \textbf{0.56}\sd{0.03} & 0.44\sd{0.02} \\

MiT-B3 & 512 & 0.54\sd{0.05} & \textbf{0.67}\sd{0.01} & \textbf{0.59}\sd{0.02} & 0.57\sd{0.02} & 0.59\sd{0.03} & 0.55\sd{0.02} & \textbf{0.45}\sd{0.02} \\

MiT-B3 & 224 & 0.56\sd{0.01} & \textbf{0.67}\sd{0.01} & 0.54\sd{0.07} & \textbf{0.59}\sd{0.01} & \textbf{0.60}\sd{0.01} & 0.55\sd{0.02} & \textbf{0.45}\sd{0.02} \\

\midrule

\multicolumn{9}{c}{\textbf{Tree cover (F1)}} \\
\midrule

MiT-B3 {\small(dice)} & 640 & 0.84\sd{0.01} & \textbf{0.92}\sd{0.00} & \textbf{0.93}\sd{0.01} & 0.92\sd{0.00} & 0.89\sd{0.00} & 0.89\sd{0.00} & \textbf{0.89}\sd{0.00} \\

MiT-B3 & 640 & 0.85\sd{0.00} & \textbf{0.92}\sd{0.02} & \textbf{0.93}\sd{0.00} & 0.92\sd{0.00} & 0.89\sd{0.00} & \textbf{0.90}\sd{0.00} & \textbf{0.89}\sd{0.00} \\

MiT-B3 & 512 & \textbf{0.86}\sd{0.02} & \textbf{0.92}\sd{0.00} & \textbf{0.93}\sd{0.00} & \textbf{0.93}\sd{0.00} & \textbf{0.90}\sd{0.01} & \textbf{0.90}\sd{0.01} & \textbf{0.89}\sd{0.00} \\

MiT-B3 & 224 & \textbf{0.86}\sd{0.01} & 0.91\sd{0.00} & \textbf{0.93}\sd{0.02} & 0.92\sd{0.00} & 0.89\sd{0.01} & \textbf{0.90}\sd{0.01} & 0.88\sd{0.00} \\

\bottomrule
\end{tabular}%
}
\end{table}

\subsection{Compute time}
We report average training time across seeds for all models evaluated in the main experiments and ablations. All models are trained on a single GPU (NVIDIA H200) under identical settings.

Among segmentation models, MiT-B3 requires $\sim$8\,h for patch size 640 and $\sim$7\,h for 512 and $\sim$7\,h for 224. MiT-B1 requires $\sim$4\,h. U-Net variants (ResNet-34 and ResNet-50) require $\sim$5\,h, while DeepLabV3+ requires $\sim$4\,h (ResNet-34) and $\sim$5\,h (ResNet-50). Mask2Former is more computationally intensive, requiring $\sim$23\,h (Tiny) and $\sim$25\,h (Small).

The foundation-model baselines fall in a similar range: DINOv2-B requires $\sim$7\,h in the single-layer setting  and $\sim$8\,h in the multi-layer setting.

Overall, training time scales with model capacity and architectural complexity, with transformer-based segmentation models incurring higher compute cost compared to lighter or frozen-backbone approaches.

\end{document}